\documentclass{article}

\usepackage[numbers, sort]{natbib}
\usepackage[preprint]{neurips_2025}
\usepackage{times}
\usepackage{latexsym}

\usepackage[T1]{fontenc}
\usepackage[utf8]{inputenc}

\usepackage{microtype}

\usepackage{inconsolata}

\usepackage{amsmath}
\usepackage{amsfonts}
\usepackage{amssymb}
\usepackage{subfigure}
\usepackage{cases}
\usepackage{dsfont}
\usepackage{bm}
\usepackage{bbm}
\usepackage{graphicx}
\usepackage{color}
\usepackage{multicol}
\usepackage{multirow}
\usepackage{wrapfig,lipsum,booktabs}
\usepackage[ruled,vlined,linesnumbered]{algorithm2e}
\usepackage{algorithmic}
\usepackage{pgfplots}
\pgfplotsset{compat=1.12} 
\usepackage{filecontents}
\usepackage{tikz}
\usepackage{xcolor}
\usepackage{colortbl}
\usepackage{xspace}
\usepackage{makecell}
\usepackage{soul}
\usepackage{subcaption}
\usetikzlibrary{calc}
\usepgfplotslibrary{groupplots}
\usetikzlibrary{angles,quotes} 
\usetikzlibrary{shapes,arrows}
\usetikzlibrary{backgrounds}
\usetikzlibrary{matrix}
\usetikzlibrary{patterns}
\usepackage{tikz-3dplot}
\usepackage{hyperref}
\usepackage{cleveref}
\usepackage{paralist}
\usepackage{cancel}
\usepackage{todonotes}
\usepackage{tabu}
\usepackage{rotating}
\usepackage{etoolbox}
\usepackage{adjustbox}
\usepackage{enumerate}
\usepackage{enumitem}
\setitemize{itemsep=0pt,topsep=0pt,parsep=0pt,partopsep=0pt}
\setenumerate{itemsep=0pt,topsep=0pt,parsep=0pt,partopsep=0pt}
\usepackage{pifont}
\usepackage{cancel}
\usepackage{lipsum}
\usepackage{listings,lstautogobble}
\usepackage{fancyvrb}
\usepackage{fvextra}
\usepackage{caption}
\usepackage{arydshln}
\usepackage{float}
\usepackage{color} 
\usepackage{colortbl}
\usepackage{pmboxdraw} % ├  └
\usepackage{wasysym}
\newcommand{\model}[1]{#1\xspace}
\newcommand{\qwen}{\model{Qwen2.5-7B-Instruct}}
\newcommand{\llamaE}{\model{Llama-3.1-8B-Instruct}}
\newcommand{\llamaT}{\model{Llama-3.2-3B-Instruct}}
\newcommand{\llamaO}{\model{Llama-3.2-1B-Instruct}}
\newcommand{\gemma}{\model{Gemma-2-2b-Instruct}}

\newcommand{\mathqwen}{\model{Qwen2.5-Math-7B-Instruct}}
\newcommand{\chatqwen}{\model{Qwen2.5-14B-Instruct}}
\newcommand{\medllama}{\model{Bio-Medical-Llama-3-8B}}
\newcommand{\finllama}{\model{Llama-3-8B-Instruct-Finance}}

\newcommand{\our}[1]{\textsc{#1}\xspace}

\newcommand{\ours}{\our{ExpertSteer}}

\newcommand{\dataset}[1]{\texttt{#1}\xspace}
\newcommand{\gsm}{\dataset{GSM8K}}
\newcommand{\nli}{\dataset{NLI}}
\newcommand{\copa}{\dataset{COPA}}

\newcommand{\mmlu}{\dataset{MMLU}}
\newcommand{\mmlumed}{\dataset{MMLU-Medical}}
\newcommand{\mmlufin}{\dataset{MMLU-Financial}}
\newcommand{\mmlumath}{\dataset{MMLU-Math}}
\newcommand{\mmluchat}{\dataset{MMLU-Humanities}}
\newcommand{\arcc}{\dataset{ARC-C}}
\newcommand{\mathtest}{\dataset{MATH500}}
\newcommand{\medmcqa}{\dataset{MedMCQA}}
\newcommand{\medqa}{\dataset{MedQA}}
\newcommand{\flare}{\dataset{Flare-cfa}}
\newcommand{\fpb}{\dataset{FPB}}
\newcommand{\salad}{\dataset{Salad}}
\newcommand{\behavior}{\dataset{Harmful Behaviors}}
\newcommand{\ultramedical}{\dataset{UltraMedical}}
\newcommand{\metamathqa}{\dataset{MetaMathQA}}
\newcommand{\lmsys}{\dataset{LMSYS-Chat-1M}}
\newcommand{\finqa}{\dataset{FINQA}}

\newcommand{\avgall}{\mu_{\textsc{all}}}

\SetCommentSty{mycommfont}

\usepackage[most]{tcolorbox}

\definecolor{paper-red}{HTML}{de5246}

\newtcbox{\clustertab}[1]{on line, box align=base, colback={#1},colframe={#1},size=fbox,arc=2pt,top=-1.5pt, bottom=-1.5pt, left=-1.5pt, right=-1.5pt, boxrule=0pt, enlarge left by=1pt}

\newcommand{\hz}{\vphantom{\parbox[c]{0.08cm}{\rule{0.08cm}{0.19cm}}}}
\definecolor{airforceblue}{rgb}{0.36, 0.54, 0.66}
\definecolor{darkgreen}{rgb}{0,.5,0} 
\definecolor{alizarin}{rgb}{0.82, 0.1, 0.26}

\newcommand{\dtcf}{%
  \colorbox{airforceblue!30}{\hz $\mathcal{XF}$}%
}
\newcommand{\dtsf}{%
  \colorbox{darkgreen!30}{\hz $\mathcal{SF}$}%
}

\newcommand{\topone}[1]{\cellcolor[HTML]{d98978}{#1}}
\newcommand{\toptwo}[1]{\cellcolor[HTML]{dfa194}{#1}}
\newcommand{\topthree}[1]{\cellcolor[HTML]{e8b3a8}{#1}}
\newcommand{\topfour}[1]{\cellcolor[HTML]{e9b7ac}{#1}}
\newcommand{\topfive}[1]{\cellcolor[HTML]{f1cdc6}{#1}}
\newcommand{\topsix}[1]{\cellcolor[HTML]{f2d5cf}{#1}}
\newcommand{\topseven}[1]{\cellcolor[HTML]{f4e5e2}{#1}}

\title{\ours: Intervening in LLMs through Expert Knowledge}
\author{%
Weixuan Wang\textsuperscript{1$\dagger$} \quad Minghao Wu\textsuperscript{2$\dagger$} \quad Barry Haddow\textsuperscript{1} \quad Alexandra Birch\textsuperscript{1} \\[1ex]
\textsuperscript{1}School of Informatics, University of Edinburgh \\
\textsuperscript{2}Monash University \\
\texttt{\{weixuan.wang, bhaddow, a.birch\}@ed.ac.uk} \\
\texttt{minghao.wu@monash.edu} 
}

\begin{document}
\maketitle

\begingroup
\renewcommand\thefootnote{}\footnotetext{
\textsuperscript{$\dagger$}~Equal contribution.
}
\endgroup

\begin{abstract}

Large Language Models (LLMs) exhibit remarkable capabilities across various tasks, yet guiding them to follow desired behaviours during inference remains a significant challenge. Activation steering offers a promising method to control the generation process of LLMs by modifying their internal activations. However, existing methods commonly intervene in the model's behaviour using steering vectors generated by the model itself, which constrains their effectiveness to that specific model and excludes the possibility of leveraging powerful \emph{external expert models} for steering. To address these limitations, we propose \textbf{\ours}, a novel approach that leverages arbitrary specialized expert models to generate steering vectors, enabling intervention in any LLMs. \ours transfers the knowledge from an expert model to a target LLM through a cohesive four-step process: first aligning representation dimensions with auto-encoders to enable cross-model transfer, then identifying intervention layer pairs based on mutual information analysis, next generating steering vectors from the expert model using Recursive Feature Machines, and finally applying these vectors on the identified layers during inference to selectively guide the target LLM without updating model parameters. We conduct comprehensive experiments using three LLMs on 15 popular benchmarks across four distinct domains. Experiments demonstrate that \ours significantly outperforms established baselines across diverse tasks at minimal cost.\footnote{\url{https://github.com/weixuan-wang123/ExpertSteer}}

\end{abstract}

\section{Introduction}
\label{sec:intro}

Large language models (LLMs) have demonstrated remarkable capabilities across diverse tasks \citep{claude35sonnet,DBLP:journals/corr/abs-2403-05530,DBLP:journals/corr/abs-2501-12948,gpt4,o1}. However, aligning these LLMs with desirable behaviour remains challenging \citep{flan,instruction,DBLP:conf/icml/WolfWALS24}. Recent research attempts to address this challenge with prompt engineering \citep{DBLP:conf/nips/BrownMRSKDNSSAA20,DBLP:conf/nips/Wei0SBIXCLZ22}, supervised fine-tuning (SFT) \citep{sft,DBLP:journals/corr/abs-2308-10792,DBLP:journals/corr/abs-2108-07258}, reinforcement learning from human feedback (RLHF)  \citep{DBLP:conf/nips/ChristianoLBMLA17,DBLP:journals/corr/abs-1909-08593,rlhf}, which typically requires extensive resources. More recently, activation steering has been proposed as an alternative for these approaches. 
This technique modifies the LLMs' internal activations at inference time, which reduces the computational cost from fine-tuning and long context and prevents the catastrophic forgetting from updating the model parameters \citep{DBLP:journals/corr/abs-2308-10248,hernandez2023inspecting,DBLP:journals/corr/abs-2310-01405,DBLP:conf/emnlp/ShinRLWS20,DBLP:conf/nips/LiuTMMHBR22}.

While activation steering has emerged as a promising approach, significant limitations hinder its broader applicability and effectiveness. Existing activation steering methods typically generate steering vectors using the model being steered itself \citep{iti,caa,DBLP:journals/corr/abs-2410-12462,DBLP:conf/icml/LiuY0Z24}. Consequently, these methods are constrained by the inherent knowledge of the LLM, which may lack the specialized expertise or deeper understanding required for certain tasks \citep{iti,caa,DBLP:journals/corr/abs-2410-12299,DBLP:conf/aaai/ChenSJLKWX24}. Additionally, the steering vectors produced by these methods are limited to influencing the behaviour of the specific model they are derived from \citep{DBLP:journals/corr/abs-2410-12299,DBLP:conf/nips/CaoZC00MC24,DBLP:journals/corr/abs-2410-01174}, making them unsuitable for cross-model steering and restricting their potential diverse applications. Moreover, as more powerful models with distinct strengths are developed, it becomes increasingly reasonable to consider leveraging these models as external resources for activation steering \citep{DBLP:conf/ssci/DongHTM23,DBLP:journals/jmlr/YouLZ0JL22,DBLP:conf/iclr/Gu0WH24}. Therefore, while activation steering holds significant promise as a flexible and scalable solution for effectively controlling LLM behaviours, its full potential remains underutilized.

To address these limitations, we introduce \textbf{\ours}, a novel activation steering framework that incorporates an arbitrary external expert model for generating steering vectors to effectively control the behaviours of any LLMs. To enable seamless cross-model steering, we first train auto-encoders \citep{auto-encoder} to align the hidden state dimensions of the expert model with those of the target LLM. Inspired by the Optimal Brain Surgeon principle  \citep{lecun1989optimal,hassibi1993optimal}, we then perform mutual information analysis on the hidden states of both models to identify the optimal subset of layer pairs for intervention. Next, we extract informative features from the identified expert layers using Recursive Feature Machines (RFMs) \cite{agop}, implemented through Kernel Ridge Regression (KRR) \citep{saunders1998ridge} and Average Gradient Outer Product (AGOP) \cite{agop}. The principal eigenvector of the resulting feature matrix for each identified expert layer is then used as the steering vector. Finally, the steering vectors are applied to the target LLM's hidden states at the identified intervention layers during inference time. By integrating auto-encoders, expert knowledge, and advanced feature extraction techniques, \ours provides an effective and efficient steering method that enables universal knowledge transfer between arbitrary pairs of models, making it a significant practical application.

To evaluate the effectiveness of \ours, we conduct extensive experiments involving three diverse LLMs and 15 widely recognized benchmarks spanning four domains: Medical, Financial, Mathematical, and General. Our study addresses two scenarios of knowledge transfer: from a domain-specific expert model to a general-purpose target LLM, and from a larger general-purpose model to a smaller general-purpose target LLM. The results show that \ours consistently outperforms previous steering methods across all tasks.

Our contributions are summarized as follows: 

\begin{itemize}

    \item We propose \ours, a novel activation steering approach that facilitates effective knowledge transfer from arbitrary expert models to any target LLMs. Leveraging techniques such as auto-encoders, mutual information analysis, and Recursive Feature Machines (RFMs), our method streamlines the steering process into four cohesive steps, extending the generalizability of activation steering and addressing the key limitations of existing approaches (see \autoref{sec:method}).

    \item We demonstrate the broad applicability and effectiveness of \ours across multiple models and tasks. Through extensive experiments with three LLMs over 15 diverse tasks spanning four domains, \ours consistently surpasses existing activation steering methods, underscoring the generalizability of \ours. (see \autoref{sec:experiment}).

    \item We provide a detailed analysis of \ours, focusing on the influence of feature extraction, expert selection, and the workflow of \ours. We also examine its computational efficiency, demonstrating \ours is highly cost-effective (see \autoref{sec:analysis}).

\end{itemize}

\section{Related Work}
\label{sec:related_work}

\paragraph{Activation Steering}

Activation steering provides a cost-effective way to steer model behaviours by directly manipulating activations during inference \citep{DBLP:journals/corr/abs-2308-10248,hernandez2023inspecting,DBLP:journals/corr/abs-2310-01405,DBLP:conf/nips/Qiu0ZKPC24}. Current research based on steering vectors which are derived from activation differences in curated parallel positive-negative pairs enables interventions to change behaviours \citep{iti,DBLP:conf/aaai/ChenSJLKWX24,DBLP:conf/icml/LiuY0Z24,DBLP:conf/nips/CaoZC00MC24,DBLP:journals/corr/abs-2410-01174} or regulate the model's inference \citep{caa,DBLP:journals/corr/abs-2308-10248,DBLP:journals/corr/abs-2410-12462,DBLP:journals/corr/abs-2410-12877} without the need for fine-tuning \citep{DBLP:conf/iclr/WeiBZGYLDDL22,DBLP:journals/corr/abs-2204-05862} or heavy in-context examples \citep{DBLP:conf/nips/BrownMRSKDNSSAA20,DBLP:conf/nips/Wei0SBIXCLZ22}. However, current methods rely on the model itself to generate steering vectors, which restricts their effectiveness to the model's inherent knowledge and exclude the potential of utilizing more powerful models for steering \citep{caa,steerbias}.

\paragraph{Knowledge Transfer}

Knowledge transfer is a well-established techniques for performance improvement, where knowledge from a source model is transferred to a target model \citep{buciluǎ2006model,DBLP:journals/corr/HintonVD15}. However, current methods, such as distillation via synthetic datasets \citep{DBLP:conf/emnlp/KimR16,DBLP:journals/corr/abs-2310-19019,DBLP:conf/acl/HsiehLYNFRKLP23,DBLP:journals/eswa/ZhouC23} and teacher-student alignment \citep{DBLP:conf/emnlp/JiaoYSJCL0L20,DBLP:journals/corr/abs-2308-02019,kd}, rely on computationally expensive fine-tuning and risk catastrophic forgetting. \citep{DBLP:journals/corr/abs-2308-08747,DBLP:journals/corr/abs-2405-09673}. This underscores the need for more efficient, parameter-free  knowledge transfer strategies.

\paragraph{Ours}

We propose \ours, a novel method that incorporates an arbitrary expert model for steering any LLMs, unlike prior approaches that generate steering vectors within the model itself \citep{iti,caa,DBLP:journals/corr/abs-2410-12299}. \ours effectively transfers the expertise to target LLMs via the steering vectors.

\begin{figure}[t]
    \centering
    \includegraphics[scale=0.6]{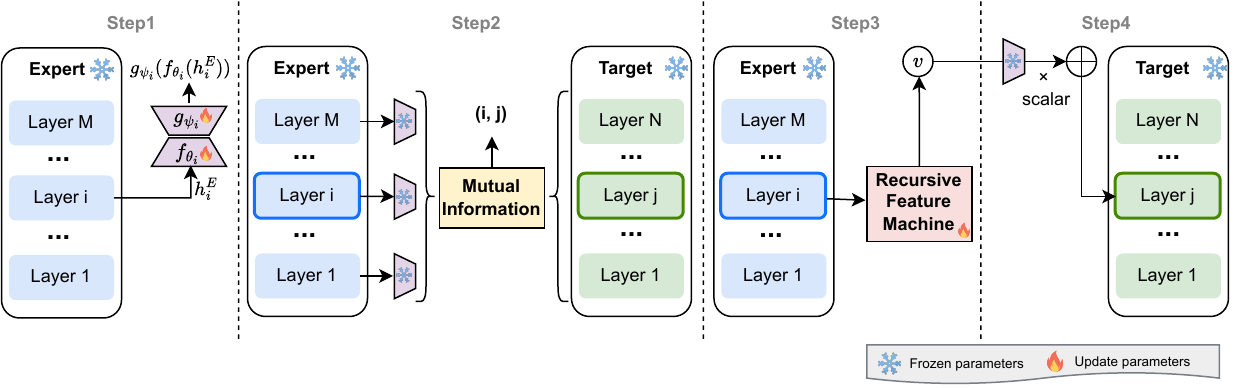}
    \caption{
    An overview of \ours, including four steps: (1) aligning the dimensionality of the expert and target models, (2) identifying the layer pairs to be intervened upon, (3) generating steering vectors from the expert model, and (4) intervening in the generation process of the target model.
    }
    \label{fig:framework}
\end{figure}

\section{\ours}
\label{sec:method}

As illustrated in \autoref{fig:framework}, we elaborate each step of \ours in this section. The first step is to align the representations between the expert model and the target model, which is detailed in \autoref{sec:method_alignment}. Next, we identify the intervention layer pairs that exhibit significant differences in their representations, as described in \autoref{sec:method_pairing}. Following this, we generate steering vectors from the expert model using Recursive Feature Machines (RFMs) in \autoref{sec:method_steering}. Finally, we apply these steering vectors to the target model during inference to enhance its performance, as outlined in \autoref{sec:method_intervention}. Furthermore, we provide implementation details in \autoref{sec:method_implementation}.

\subsection{Representation Alignment}
\label{sec:method_alignment}

A significant challenge in transferring knowledge between different models is their architectural differences, particularly the varying dimensions of hidden states across models. To address this, we introduce a representation alignment procedure that unifies the feature spaces of the expert model and the target model. For each layer $i$ in the expert model with hidden states $h_i^E \in \mathbb{R}^{d_E}$, we train a dedicated auto-encoder consisting of an encoder $f_{\theta_i}: \mathbb{R}^{d_E} \rightarrow \mathbb{R}^{d_T}$ and a decoder $g_{\phi_i}: \mathbb{R}^{d_T} \rightarrow \mathbb{R}^{d_E}$, where $d_E$ and $d_T$ represent the hidden dimensions of the expert and target models, respectively. Here, both the encoder and decoder are implemented as one affine linear layer. The auto-encoder is optimized using a reconstruction loss function:
\begin{equation}
    \mathcal{L}_{\text{recon}} = \frac{1}{K} \sum_{k=1}^{K} \| h_{i,k}^E - g_{\phi_i}(f_{\theta_i}(h_{i,k}^E)) \|_2^2
\end{equation}
where $K$ is the number of training examples. This loss ensures that the encoder-decoder pair can effectively compress and expand the expert model's representations while preserving essential information. The trained encoder $f_{\theta_i}$ serves as a bridge between the expert and target feature spaces, enabling us to project the expert's hidden states into a form compatible with the target model.

\subsection{Intervention Layer Pairing}
\label{sec:method_pairing}

After aligning the representations between the expert and target models, the next step is to identify the layer-wise pairing relationship between the two models. Inspired by the Optimal Brain Surgeon (OBS) principle, which emphasizes that effective neural network modifications should be both selective and minimal \citep{lecun1989optimal,hassibi1993optimal}, we intervene in only a subset of the target model's layers. This selective strategy maximizes the potential benefits of the intervention while minimizing the risk of introducing noise.

Mutual information (MI) quantifies the amount of information obtained about one random variable through observing another random variable, making it an ideal metric for measuring representation alignment between two models. Hence, we conduct a layer-wise MI analysis to identify the layer pairs for steering. For each layer pair $(i, j)$, where $i$ refers to a layer in the expert model and $j$ refers to a layer in the target model, we follow \cite{zhelezniak-etal-2020-estimating} to estimate the MI between hidden states: 
\begin{equation}
    \text{MI}(i, j) = \frac{1}{K} \sum_{k=1}^{K} \mathbb{I}(f_{\theta_i}(h_{i,k}^E); h_{j,k}^T), \quad \text{where} \quad \mathbb{I}(X;Y) = \int\int p(x,y) \log\frac{p(x,y)}{p(x)p(y)} dx dy
\end{equation}
Here, $\mathbb{I}(\cdot;\cdot)$ denotes the mutual information operator, measuring the reduction in uncertainty about $Y$ when $X$ is known, and $K$ is the number of examples used for estimate MI. For $k$-th example, the expert’s hidden states at $i$-th layer $h_{i,k}^E$ are mapped to the target’s dimensionality by the encoder $f_{\theta_i}$, and $h_{j,k}^T$ represents the hidden states of the target model at layer $j$. 
Lower MI indicates a greater disparity between the expert layer and the target layer, implying that the representation at the target layer potentially lacks the expert's knowledge. This suggests a greater need for intervention. Conversely, higher MI implies that the target model's representation is already well-aligned with the expert's, thereby reducing the necessity for intervention.

Subsequently, we select intervention points where knowledge transfer would be most beneficial. Specifically, we compute the MI for all layer pairs $(i, j)$ and select the top-$P$ pairs with the lowest values. These low-MI pairs represent areas where the target's representations diverge most significantly from the expert's knowledge, making them better candidates for intervention.

\subsection{Steering Vector Generation}
\label{sec:method_steering}

After identifying the intervention layer pairs, we need to generate steering vectors that encode the expert model's knowledge. To this end, we employ Recursive Feature Machines (RFMs) \cite{agop} to extract the most informative features from the expert model's hidden states.
In our approach, the RFMs algorithm employs two key components: Kernel Ridge Regression (KRR) \citep{saunders1998ridge} and the Adaptive Gradient Optimal Perturbation (AGOP) matrix \cite{agop}. 
The KRR model learns to distinguish between hidden states given by inputs from different sources by binary classification, while the AGOP matrix captures the feature importance by analysing gradients of the KRR model.

For each selected expert model layer $i$, we gather hidden states $H_i = [h_{i,1}^E, h_{i,2}^E, \ldots, h_{i,K}^E] \in \mathbb{R}^{K \times d_E}$ from $K$ training examples. Each example is assigned a binary label with One-vs-Rest strategy: positive (1) for examples that align with the expert's knowledge, and negative (0) for examples that do not. For instance, when using a medical LLM as the expert, examples related to medical topics are labelled as positive, while examples unrelated to the medical domain are labelled as negative. 

\begin{algorithm}[h]
\caption{Recursive Feature Machines (RFMs)}
\label{alg:RFMs}
\SetKwInOut{Input}{Input}
\SetKwInOut{Output}{Output}
\Input{Training data $H_i = [h_{i,1}^E, h_{i,2}^E, \ldots, h_{i,K}^E] \in \mathbb{R}^{K \times d_E}$; binary labels $Y = [y_1, y_2, \ldots, y_K] \in \mathbb{R}^K$; the number of iterations $\tau$; the bandwidth parameter $\sigma$; the number of training examples $K$.}
\Output{Feature importance matrix $M_i^{\tau}$}
\SetAlgoLined

$M_i^0 \leftarrow I_{d_E}$ \tcp*{Initialize feature importance matrix}

\For{$t = 0$ to $\tau-1$}{
    $\mathbb{K}^t(h_{i,k}^E, z) \leftarrow \exp\left(-\frac{1}{\sigma}(h_{i,k}^E - z)^{\top}\mathcal{M}_i^t(h_{i,k}^E - z)\right)$ \tcp*{Update kernel function} \label{line:laplace}
    $\beta_t \leftarrow (\mathbb{K}^t(H_i, H_i))^{-1} Y$ \tcp*{Solve $\beta_t$ for the predictor $\pi^t(z) = \mathbb{K}^t(H_i,z) \beta_t$} \label{line:krr}
    $\mathcal{M}_i^{t+1} \leftarrow \frac{1}{K}\sum_{k=1}^K \nabla_{h_{i,k}^E} \pi^t(h_{i,k}^E) \cdot (\nabla_{h_{i,k}^E}\pi^t(h_{i,k}^E))^{\top} $ \tcp*{Compute AGOP matrix} \label{line:agop}
}

\end{algorithm}

As detailed in \autoref{alg:RFMs}, in each iteration $t$ of the RFMs, we first update the Mahalanobis Laplace Kernel function $\mathbb{K}^t$ using the current feature importance matrix $\mathcal{M}_i^t$ (\autoref{line:laplace}), where $z$ in $\mathbb{K}^t$ indicates an arbitrary hidden state from $H_i$. This adaptive kernel measures the similarity between hidden states while accounting for their relevance to domain distinction. We then solve for coefficients $\beta_t$ using KRR, which optimizes the predictor $\pi^t(z) = \mathbb{K}^t(H_i, z) \beta_t$ to classify representations by domain (\autoref{line:krr}). Finally, we update the feature importance matrix $\mathcal{M}_i^{t+1}$ by computing AGOP, which averages the outer products of gradients across all training examples (\autoref{line:agop}). After $\tau$ iterations, the final matrix $\mathcal{M}_i^{\tau}$ captures directions in the feature space that most reflect desired knowledge.

To extract the steering vector from this feature importance matrix, we perform eigenvalue decomposition on $\mathcal{M}_i^{\tau} = U\Lambda U^{\top}$, where $\Lambda = \text{diag}(\lambda_1, \lambda_2, \ldots, \lambda_{d_E})$ are the eigenvalues (sorted in descending order) and $U = [u_1, u_2, \ldots, u_{d_E}]$ are the corresponding eigenvectors. The eigenvector $u_1$ associated with the largest eigenvalue $\lambda_1$ represents the direction of maximum variance in the feature space, capturing the most desired knowledge. We define $u_1$ as the steering vector $\nu_i$ for the $i$-th layer. This approach ensures that our intervention targets the most salient aspects of the expert model's knowledge, maximizing the effectiveness of the knowledge transfer.

\subsection{Expertise Intervention}
\label{sec:method_intervention}

In the final step, we transfer the expert knowledge distilled in the steering vectors to the target model by intervening at the $P$ most impactful layer pairs $(i, j)$ identified previously. Since the expert and target models may have different hidden dimensions ($d_E$ and $d_T$), we ensure compatibility by leveraging the encoder $f_{\theta_i}(\cdot)$ from the trained auto-encoder (see \autoref{sec:method_alignment}). This encoder projects the expert's steering vector $\nu_i \in \mathbb{R}^{d_E}$ into the target model's feature space $\mathbb{R}^{d_T}$ when necessary. Formally, for each selected layer pair $(i, j)$, we update the hidden state $h_j^T$ of the target model:
\begin{align}
    \hat{h}_j^T = \begin{cases} 
        h_j^T + \varepsilon \cdot f_{\theta_i} (\nu_i) & \text{if } d_E \neq d_T \\
        h_j^T + \varepsilon \cdot \nu_i & \text{if } d_E = d_T
    \end{cases}
    \label{eq:intervention}
\end{align}
where $\varepsilon$ is a scaling factor controlling the strength of the intervention. The modified hidden state $\hat{h}_j^T$ is then propagated through the remaining layers of the target model to produce the final output.

\subsection{Implementation Details}
\label{sec:method_implementation}

Our method introduces two hyperparameters: $P \in \mathbb{N}^+$, specifying the number of top layer pairs selected for intervention, and $\varepsilon \in \mathbb{R}^+$, controlling the strength of the intervention. In our experiments, we explore $P$ values ranging from 1 to 10, and $\varepsilon$ values in $\{1, 2, 4, 6, 8, 10, 12, 14, 16\}$. Following \citep{iti,caa,DBLP:journals/corr/abs-2410-12299}, we perform a hyperparameter sweep to empirically determine the optimal settings on a small development set, which are subsequently utilized during the final evaluation on the test set. 

We use 2,000 random examples to train the auto-encoders in \autoref{sec:method_alignment}. Then, we leverage 500 random examples to identify the intervention pairs in \autoref{sec:method_pairing}. And, we sample 2,000 positive examples and 2,000 negative examples to train RFMs in \autoref{sec:method_steering}. More details are in \autoref{sec:appendix_training}.

\begin{table}[b]
\centering
\setlength{\tabcolsep}{2pt}
\small
\caption{The datasets used for training and evaluation, and the expert models utilized in this work.}
\begin{tabular}{@{}llp{4.5cm}p{0.1cm}l@{}}
\toprule
             & Training  Datasets                        & Evaluation Datasets &  & Expert Model \\ \hline
\cellcolor{gray!15}Medical      & \cellcolor{gray!15}\ultramedical\citep{ultramedical} & \cellcolor{gray!15}\medqa \citep{medqa}, \medmcqa \citep{medmcqa}, \mmlumed \citep{mmlu}  & \cellcolor{gray!15} & \cellcolor{gray!15}\medllama \citep{ContactDoctor}                                          \\ 
Financial    & \finqa \citep{finqa}              & \fpb \citep{fpb}, \flare \citep{flare}, \mmlufin \citep{mmlu}       &  &  \finllama \citep{finllama} \\
\cellcolor{gray!15}Mathematical & \cellcolor{gray!15}\metamathqa \citep{metamathqa}    & \cellcolor{gray!15}\gsm \citep{gsm8k}, \mathtest \citep{math}, \mmlumath \citep{mmlu}      & \cellcolor{gray!15}   & \cellcolor{gray!15}\mathqwen \citep{qwenmath}                                            \\
General      & \lmsys \citep{lmsys}              & \copa \citep{copa}, \nli \citep{xnli}, \arcc \citep{arc}, \mmluchat \citep{mmlu}, \salad \citep{salad}, \behavior \citep{harmbehavior}    &    & \chatqwen \citep{qwen}                           \\
\bottomrule
\end{tabular}
\label{tab:datasets}
\end{table}

\section{Experiments}
\label{sec:experiment}
\begin{table}[t] \small
\centering
\setlength{\tabcolsep}{3.3pt}
\caption{
Results on the Medical, Financial and Mathematical domains with \llamaE, \qwen, and \gemma target models across \colorbox[HTML]{E3D4E7}{discriminative tasks} and \colorbox[HTML]{FFF3CE}{generative tasks}. The expert models are \medllama, \finllama and \mathqwen. Same-Family (\dtsf) and Cross-Family (\dtcf) indicates that if the expert and target model belong to the same model family. The \colorbox[HTML]{d98978}{\textbf{best overall}} results are highlighted.
}
\begin{tabular}{lrcccrcccrccc}
\toprule
 & \multicolumn{4}{c}{Medical}       & \multicolumn{4}{c}{Financial}        & \multicolumn{4}{c}{Mathematical}           \\ \cmidrule(rl){2-5} \cmidrule(rl){6-9} \cmidrule(rl){10-13}
\multicolumn{1}{c}{} & $\avgall$  & \cellcolor[HTML]{E3D4E7}\dataset{MedQA} & \cellcolor[HTML]{E3D4E7}\begin{tabular}[c]{@{}c@{}} \dataset{Med}\\ \dataset{MCQA} \end{tabular} & \cellcolor[HTML]{E3D4E7}\begin{tabular}[c]{@{}c@{}}\dataset{MMLU}\\ \dataset{Med.} \end{tabular} & $\avgall$  & \cellcolor[HTML]{E3D4E7}\fpb   & \cellcolor[HTML]{E3D4E7}\begin{tabular}[c]{@{}c@{}} \dataset{Flare}\\ \dataset{-cfa}\end{tabular} & \cellcolor[HTML]{E3D4E7}\begin{tabular}[c]{@{}c@{}} \dataset{MMLU} \\ \dataset{Fin.}\end{tabular} & $\avgall$  & \cellcolor[HTML]{E3D4E7}\begin{tabular}[c]{@{}c@{}} \dataset{MMLU} \\ \dataset{Math}\end{tabular} & \cellcolor[HTML]{FFF3CE}\gsm & \cellcolor[HTML]{FFF3CE}\begin{tabular}[c]{@{}c@{}}\dataset{MATH}\\ \dataset{500}\end{tabular} \\ \midrule
Expert Model            & \multicolumn{1}{r}{76.61}   & 73.85  & 69.01         & 86.96          & \multicolumn{1}{r}{60.01}   & 64.34& 59.49           & 56.20          & \multicolumn{1}{r}{58.55} & 25.09          & 91.60  & 58.95         \\
\midrule
\multicolumn{13}{c}{\cellcolor{gray!15}\textbf{\llamaE}} \\
Baseline           & \topsix52.00& 45.60  & 49.40         & 60.99          & \topseven45.98 & 41.55& 48.14           & 48.26          & \topfour51.43        & 25.48          & 86.80  & 42.00         \\ 
\arrayrulecolor{gray!50} \cdashline{1-1}
\rule{0pt}{2ex}SFT & \toptwo56.44 & 53.50  & \textbf{51.35}& 64.46          & \toptwo55.73       & 54.84& 56.00           & 56.35          & \topfive46.17         & 22.51          & 80.00  & 36.00         \\
KD  & \topthree56.06              & 53.56  & 48.98         & 65.65          & \textbf{\topone56.16}       & 55.11& \textbf{56.68}  & \textbf{56.70} & \topsix44.91         & 21.32          & 78.80  & 34.60         \\
\arrayrulecolor{gray!50} \cdashline{1-1}
\rule{0pt}{2ex}ITI & \topfour54.34& 50.71  & 50.11         & 62.20          & \topfive49.01& 47.80& 49.91           & 49.31          & \toptwo52.86          & 29.17          & 87.00  & 42.40         \\
CAA & \topseven46.60 & 38.86  & 45.72         & 55.22          & \topsix47.39& 50.21& 46.23           & 45.72          & \topseven34.83          & 26.10          & 55.20  & 23.20         \\
SADI  & \topfive53.51           & 50.51& 47.02      & 62.99      & \topfour49.61     & 51.96& 47.00        & 49.87      & \topthree52.62     & 28.07       & 87.00& 42.80      \\
\multirow{2}{*}{\ours} & \textbf{\topone56.98}       & \textbf{53.59}            & 50.66         & \textbf{66.71} & \topthree51.49              & \textbf{55.21}          & 48.92           & 50.35          & \textbf{\topone54.92} & \textbf{31.95} & \textbf{88.40}            & \textbf{44.40}\\
& \multicolumn{4}{c}{\dtsf}       & \multicolumn{4}{c}{\dtsf}       & \multicolumn{4}{c}{\dtcf} \\
\arrayrulecolor{gray!20}\midrule
\multicolumn{13}{c}{\cellcolor{gray!15}\textbf{\qwen}}  \\
Baseline           & \topsix49.65& 41.20  & 46.25         & 61.50          & \topsix65.53& 76.23& 57.88           & 62.49          & \topthree55.05        & 26.75          & 89.20  & 49.20         \\
\arrayrulecolor{gray!50} \cdashline{1-1}
\rule{0pt}{2ex}SFT & \textbf{\topone55.55}       & 45.30  & \textbf{51.02}& \textbf{70.32} & \toptwo67.73 & 74.59& 59.78           & 68.83          & \topfive53.48         & 30.04          & 83.20  & 47.20         \\
KD  & \topthree53.20              & 43.68  & 47.68         & 68.23          & \topthree66.44              & 76.59& 58.36           & 64.37          & \toptwo56.88          & 31.03          & \textbf{90.80}            & 48.80         \\
\arrayrulecolor{gray!50} \cdashline{1-1}
\rule{0pt}{2ex}ITI & \topseven49.55 & 41.46  & 45.78         & 61.40          & \topseven60.25 & 76.42& 42.47           & 61.86          & \topsix49.85         & 11.14          & 90.00  & 48.40         \\
CAA & \topfive50.04& 41.46  & 46.18         & 62.48          & \topfive65.65& 76.63& 56.54           & 63.79          & \topseven42.48          & 11.85          & 81.20  & 34.40         \\
SADI  & \topfour50.38           & 42.34& 45.72      & 63.09      & \topfour66.24     & 76.91& 57.95        & 63.88      & \topfour52.95     & 22.05       & 88.80& 48.00      \\
\multirow{2}{*}{\ours} & \toptwo54.03 & \textbf{45.98}            & 48.57         & 67.53          & \textbf{\topone70.87}       & \textbf{78.40}          & \textbf{63.23}  & \textbf{71.00} & \textbf{\topone57.26} & \textbf{31.17} & \textbf{90.80}            & \textbf{49.80}\\
& \multicolumn{4}{c}{\dtcf}       & \multicolumn{4}{c}{\dtcf}       & \multicolumn{4}{c}{\dtsf} \\
\arrayrulecolor{gray!20}\midrule
\multicolumn{13}{c}{\cellcolor{gray!15}\textbf{\gemma}} \\
Baseline           & \topfive31.17& 28.63  & 33.06         & 31.81          & \topfive37.15& 47.27& 36.00           & 28.17          & \toptwo37.94          & 23.03          & 67.60  & 23.20         \\
\arrayrulecolor{gray!50} \cdashline{1-1}
\rule{0pt}{2ex}SFT & \textbf{\topone40.60}       & \textbf{37.13}            & 32.74         & \textbf{51.93} & \textbf{\topone48.76}       & \textbf{53.29}          & \textbf{46.67}  & \textbf{46.33} & \topsix35.11         & 23.33          & 57.60  & 24.40         \\
KD  & \toptwo39.79 & 35.80  & 33.19         & 50.39          & \toptwo46.54 & 50.71& 45.56           & 43.33          & \topseven33.99          & 21.17          & 56.80  & 24.00         \\
\arrayrulecolor{gray!50} \cdashline{1-1}
\rule{0pt}{2ex}ITI & \topfour31.23& 28.78  & 33.12         & 31.80          & \topfour37.61& 48.25& 36.09           & 28.50          & \topfour36.75        & 21.46          & 68.00  & 20.80         \\
CAA & \topseven30.65 & 28.17  & 32.65         & 31.14          & \topfive37.15 & 46.85& 36.59           & 28.02          & \topfive35.75         & 22.85          & 61.20  & 23.20         \\
SADI  & \topsix30.99           & 28.99& 32.03      & 31.96      & \topfour38.41     & 49.90& 36.63        & 28.71      & \topthree37.62     & 22.05       & 67.60& 23.20      \\
\multirow{2}{*}{\ours} & \topthree32.21              & 29.39  & \textbf{33.37}& 33.87          & \topthree39.47              & 51.21& 37.40           & 29.80          & \textbf{\topone39.28} & \textbf{24.24} & \textbf{68.40}            & \textbf{25.20}\\
& \multicolumn{4}{c}{\dtcf}       & \multicolumn{4}{c}{\dtcf}       & \multicolumn{4}{c}{\dtcf} \\        
\arrayrulecolor{black}\bottomrule
\end{tabular}
\label{tab:main}
\end{table}

\subsection{Experimental Setup}
\label{sec:experiment_setting}

\paragraph{Datasets and Models}
We conduct our experiments across four domains: Medical, Financial, Mathematical, General and present the datasets in \autoref{tab:datasets}. We denote the overall performance within one domain as $\avgall$, which is the macro-average of the tasks. 
We apply \ours to three target models from different families and sizes: \llamaE \citep{llama}, \qwen \citep{qwen}, and \gemma \citep{gemma}. The expert models used in the experiments are shown in \autoref{tab:datasets}.

\paragraph{Baselines}

We compare \ours with several fine-tuning baselines: standard Supervised Fine-Tuning (SFT) and Knowledge Distillation (KD) \citep{kd}, and the state-of-the-art steering baselines, including Inference-Time Intervention (ITI) \citep{iti},Contrastive Activation Addition (CAA) \citep{caa}, and Semantic-Adaptive Dynamic Intervention (SADI) \citep{DBLP:journals/corr/abs-2410-12299}. More details are shown in \autoref{sec:appendix_baselines}.

\subsection{Overall Performance}
\label{sec:experiment_results}

\paragraph{\ours effectively transfers domain-specific knowledge and significantly enhances model performance on both discriminative and generative tasks.}

As shown in \autoref{tab:main}, \ours consistently boosts performance across three target models and three domains, outperforming other intervention methods and matching or surpassing fully fine-tuned approaches like SFT and KD. In the Medical and Financial domains, it provides average gains of +4.98 for \llamaE and +5.34 for \qwen. Furthermore, \ours consistently outperforms SFT and KD in the Mathematical domain, demonstrating its superior efficiency for highly complex tasks. Even when target models occasionally outperforms expert models, \ours discovers additional knowledge through steering vectors. For example, on the \fpb benchmark, the \qwen baseline and expert models achieve scores of 76.23 and 64.34, respectively, while \ours achieves 78.40. This underscores the effectiveness of \ours in transferring expertise.

\paragraph{\ours consistently excels in both same-family and cross-family settings.}
In practice, expert and target models are likely to come from different families. Hence, we evaluate \ours under both same-family (\dtsf) and cross-family (\dtcf) settings, where same-family  indicates that the expert model and the target model belong to the same model family, while cross-family indicates that they belong to different families. As shown in \autoref{tab:main}, \ours consistently outperforms the baseline in both settings, showing gains of +4.98, +5.51, and +1.34 in three domains for same-family, and +4.38 (Medical) and +5.34 (Financial) in cross-family settings using \qwen as the target. These results confirm that \ours effectively extracts and transfers expertise despite model disparities, demonstrating its applicability and  generalizability.

\begin{wraptable}[31]{t}{9.5cm}
\centering
\small
\setlength{\tabcolsep}{2pt}
\caption{
General domain performance on the NLU tasks and Safety tasks with \llamaE, \qwen, and \gemma target models across \colorbox[HTML]{E3D4E7}{discriminative tasks} and \colorbox[HTML]{FFF3CE}{generative tasks}. The expert model is \chatqwen. 
}
\begin{tabular}{lcccccccc}
\toprule
& \multicolumn{5}{c}{NLU}      & \multicolumn{3}{c}{Safety}   \\ \cmidrule(rl){2-6} \cmidrule(rl){7-9}
& $\avgall$ &  \cellcolor[HTML]{E3D4E7}\copa     &  \cellcolor[HTML]{E3D4E7}\nli      &  \cellcolor[HTML]{E3D4E7}\arcc     &  \cellcolor[HTML]{E3D4E7}\begin{tabular}[c]{@{}l@{}}\dataset{MMLU}\\ \dataset{Hum.} \end{tabular} & $\avgall$ &  \cellcolor[HTML]{FFF3CE}\salad & \cellcolor[HTML]{FFF3CE}\begin{tabular}[c]{@{}l@{}} \dataset{Harm}\\ \dataset{Behav.} \end{tabular} \\ \midrule
Expert Model    & 82.42       & 96.60 & 75.66 & 82.72 & 74.71      & 83.20       & 78.40 & 88.00\\
 \midrule
\multicolumn{9}{c}{\cellcolor{gray!15}\textbf{\llamaE} \quad \dtcf}   \\
Baseline   & \topfive64.68        & 74.01& 57.87& 67.39& 59.45& \topsix65.20        & 57.20& 73.20     \\
ITI        & \toptwo67.01        & 81.75& 57.82& \textbf{68.97} & 59.49& \toptwo72.60        & \textbf{72.00} & 73.20     \\
CAA        & \topsix63.11        & 80.90& 50.47& 64.13& 56.93& \topthree72.40        & 71.60& 73.20     \\
SADI     & \topthree65.32   & 81.36    & 57.98    & 64.93    & 57.00& \topfive72.20   & 71.60    & 72.80     \\
\ours & \textbf{\topone68.45}         & \textbf{83.47} & \textbf{61.36} & 68.34& \textbf{60.60} & \textbf{\topone72.80}         & \textbf{72.00} & \textbf{73.60}      \\
\arrayrulecolor{gray!20}\midrule
\multicolumn{9}{c}{\cellcolor{gray!15}\textbf{\qwen} \quad \dtsf }       \\
Baseline  & \topsix72.51       & 82.07& 71.00& 73.54& 63.41      & \topsix77.60       & 72.80& 82.40\\
ITI       & \topfive72.74       & 82.03& 73.63& 72.95& 62.34      & \toptwo80.20       & 75.60 & 84.80\\
CAA       & \topthree73.66       & 84.20& 73.25& 74.26& 62.94      & \textbf{\topone82.40}         & 74.80& \textbf{90.00}  \\
SADI     & \toptwo74.08   & 85.37      & 73.99      & 73.98      & 62.98& \topthree81.30   & \textbf{76.00}                     & 86.60     \\
\ours & \textbf{\topone77.53}         & \textbf{88.23} & \textbf{77.20} & \textbf{78.24} & \textbf{66.44}        & \topfive79.20       & 74.80& 83.60\\
 \arrayrulecolor{gray!20}\midrule
\multicolumn{9}{c}{\cellcolor{gray!15}\textbf{\gemma} \quad \dtcf}       \\
Baseline  & \toptwo46.67       & 72.32& 41.82& 38.17& 34.36      & \topsix78.60       & 74.80& 82.40\\
ITI       & \topthree46.38       & 73.34& 40.21& 37.85& 34.11      & \topfive80.80       & 77.60& 84.00\\
CAA       & \topsix45.73       & 69.75& 42.38& 37.40& 33.39      & \textbf{\topone81.30}         & 78.00& \textbf{84.60}  \\
SADI     & \topfive45.76   & 71.14      & 40.50      & 37.03      & 34.36& \topthree81.10   & 78.00      & 84.20     \\
\ours & \textbf{\topone48.35}         & \textbf{75.57} & \textbf{44.10} & \textbf{39.11} & \textbf{34.63}        & \textbf{\topone81.30}         & \textbf{78.40} & 84.20\\
\arrayrulecolor{black}\bottomrule
\end{tabular}

\label{tab:general}
\end{wraptable}
\paragraph{\ours can also improve the model performance when the expert and target models share the same domain.}
While \ours effectively transfers knowledge across domains, we also investigate its potential to enhance model performance when both the expert and target models belong to the same domain. To this end, we leverage the general-purpose \chatqwen as the expert model and present the results in \autoref{tab:general}. The results demonstrate that \ours consistently outperforms other steering methods on both natural language understanding (NLU) and safety tasks. Unlike prior steering methods, which are often constrained by the model's inherent capabilities, \ours effectively leverages the strengths of more powerful models, thereby unlocking their full potential. These findings highlight the versatility and effectiveness of \ours in both cross-domain and same-domain scenarios.

% \begin{table}[t]
\begin{wraptable}[15]{t}{7.2cm}
\centering
\small
\setlength{\tabcolsep}{0.8pt}
\caption{
Chinese Performance on two target models with expert model \model{Llama3.1-8B-Chinese-Chat}. \dataset{xsc} represents \dataset{XStoryCloze}.
}
\begin{tabular}{lcccccc}
\toprule
& $\avgall$& \begin{tabular}[c]{@{}c@{}} \dataset{XCOPA} \\ \dataset{-zh}\end{tabular} & \begin{tabular}[c]{@{}c@{}}\dataset{XNLI}\\ \dataset{-zh}\end{tabular} & \begin{tabular}[c]{@{}c@{}} \dataset{xsc}\\ \dataset{-zh}\end{tabular} & \begin{tabular}[c]{@{}c@{}}\dataset{Flores}\\ \dataset{-en2zh}\end{tabular} & \begin{tabular}[c]{@{}c@{}}\dataset{Flores}\\ \dataset{-zh2en}\end{tabular} \\ \midrule
Expert Model & 57.58     & 87.13       & 60.14      & 87.86        & 32.79           & 19.96           \\ \midrule
\multicolumn{7}{c}{\cellcolor{gray!15}\llamaE} \\
Baseline & 49.56 & 77.58            & 49.17           & 76.10             & 26.36                & 18.58                \\
\ours    & \textbf{50.98} & \textbf{78.32}            & \textbf{49.63}           & \textbf{76.39}            & \textbf{31.11}               & \textbf{19.46}               \\
\multicolumn{7}{c}{\cellcolor{gray!15}\qwen}  \\
Baseline     & 58.22     & 79.60       & 63.39      & 93.25        & 34.95           & 19.90           \\
\ours        & \textbf{62.82}     & \textbf{91.69 }      & \textbf{71.90}      & \textbf{94.85}        & \textbf{35.05}           & \textbf{20.62}          \\
\bottomrule
\end{tabular}

\label{tab:language}
% \end{table}
\end{wraptable}

\paragraph{\ours effectively transfers linguistic expertise.}

While our primary experiments focus on English, we extend \ours to other languages to demonstrate its broader applicability. Specifically, we evaluate \ours on Chinese datasets: \dataset{XCOPA-zh}~\citep{xcopa}, \dataset{XNLI-zh}~\citep{xnli}, \dataset{XStoryCloze-zh}~\citep{xstorycloze}, \dataset{Flores-en2zh}, and \dataset{Flores-zh2en}~\citep{flores}, using the expert model \model{Llama3.1-8B-Chinese-Chat}~\citep{zhllama}. The steering vector is extracted from 2,000 items randomly selected from the Chinese News Commentary dataset. Results in \autoref{tab:language} show consistent performance gains for both \llamaE and \qwen, confirming that the effectiveness of \ours extends beyond English.

\section{Analysis}
\label{sec:analysis}
% % \begin{table}[tt]

\begin{wraptable}[16]{t}{5.3cm}

\small
\centering
\setlength{\tabcolsep}{3pt}
\caption{
Comparison between different feature extraction methods.
}
\begin{tabular}{lcccc}
\toprule
& \dataset{MedQA}     & \begin{tabular}[c]{@{}c@{}}\dataset{MMLU}\\ \dataset{Med.} \end{tabular} & \copa     & \begin{tabular}[c]{@{}c@{}}\dataset{MMLU}\\ \dataset{Hum.}\end{tabular} \\ \midrule
\multicolumn{5}{c}{\cellcolor{gray!15}\llamaE}                                     \\
Baseline & 45.60 & 60.99  & 74.01 & 59.45 \\
\multicolumn{5}{l}{\ours} \\
├ MD       & 42.56 & 59.11  & 81.19 & 59.88 \\
├ PCA      & 42.58 & 59.17  & 83.10 & 59.89 \\
└ RFMs  & \textbf{53.59} & \textbf{66.71}  & \textbf{83.47} &\textbf{ 60.60} \\
\multicolumn{5}{c}{\cellcolor{gray!15}\qwen}                                      \\
Baseline & 41.20 & 61.50  & 82.07 & 63.41 \\
\multicolumn{5}{l}{\ours} \\
├ MD       & 43.18 & 64.97  & 82.01 & 64.10 \\
├ PCA      & 44.52 & 65.30  & 86.01 & 64.59 \\
└ RFMs  & \textbf{45.98} & \textbf{67.53}  & \textbf{88.23} & \textbf{66.44}
    \\
\bottomrule
\end{tabular}

\label{tab:feature}
% % \end{table}
\end{wraptable}

In this section, we firstly conduct ablation studies to analyse \ours in \autoref{sec:analysis_ablation}, including the impact of feature extraction methods, the choice of the expert models, and the order of operations. We examine the computational efficiency and explore how the foundation models, model sizes affect performance of \ours in \autoref{sec:analysis_discussion}. We present results on hyperparameter, kernel types in \autoref{sec:appendix-hyperparameter} and \autoref{sec:appendix_kernel}.

\subsection{Ablation Studies}
\label{sec:analysis_ablation}

\paragraph{RFMs excel in feature extraction.}

Unlike linear activation steering methods, \ours uses RFMs with a non-linear kernel to extract steering vectors. To validate effectiveness of RFMs, we compare RFMs with linear approaches, such as mean difference (MD) and Principal Component Analysis (PCA) on the medical and general tasks. As shown in \autoref{tab:feature}, \ours with RFMs consistently outperforms those with MD or PCA across all evaluations. Among linear methods, PCA often exceeds MD by capturing higher-dimensional variance, while MD only considers first-order statistical differences between domains. More results are presented in \autoref{sec:appendix_feature}.

\begin{figure}[t]
  \centering
  \begin{minipage}{0.58\linewidth}
    \centering
    \includegraphics[width=\linewidth]{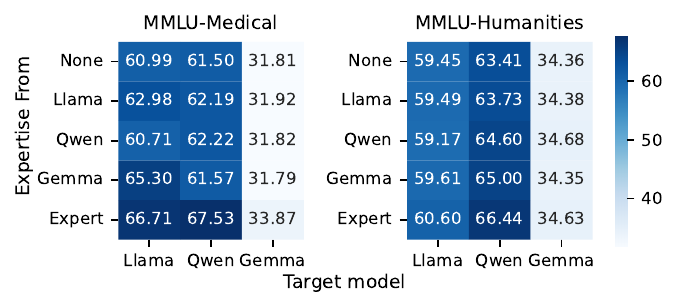}
    \caption{The selection of model for generating steering vectors. ``None'' indicates no expert is used. ``Expert'' represents the models in \autoref{tab:datasets}. ``Llama'', ``Qwen'', ``Gemma'' represent \llamaE, \qwen, and \gemma, respectively.}
    \label{fig:modelA}
  \end{minipage}
  \hfill
  \begin{minipage}{0.4\linewidth}
    \centering
    \includegraphics[width=0.9\linewidth]{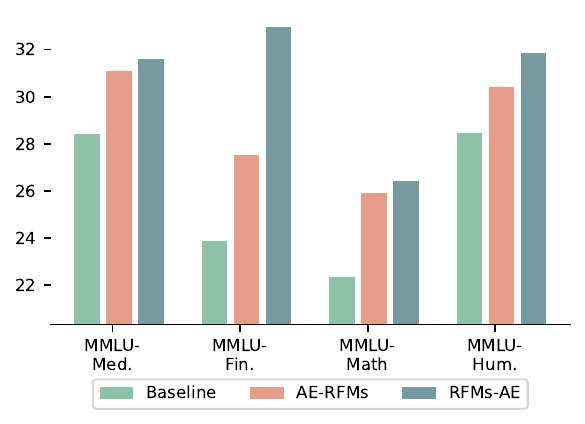}
    \caption{Comparison of \emph{RFMs-AE} and \emph{AE-RFMs} using \llamaO: \emph{RFMs-AE} extracts features before aligning dimensions, yet \emph{AE-RFMs} aligns dimensions before feature extraction.}
    \label{fig:order-llama1b}
  \end{minipage}
\end{figure}

\paragraph{The choice of the expert model is essential for activation steering.}
 
Expert model selection is vital for \ours. As illustrated in \autoref{fig:modelA}, we evaluate the performance of \ours using steering vectors generated by various models, including general-purpose models (Llama, Qwen, and Gemma) and expert models. We observe that steering vectors from experts significantly outperform those from general-purpose models, as they better capture most salient desired features. For instance, applying \llamaE on itself yields only a slight improvement (62.98 versus baseline 60.99 on \mmlumed), whereas expert models deliver a substantial boost (e.g., 66.23). Furthermore, we observe similar patterns on the \mmluchat in \autoref{fig:modelA}. These findings highlight the limitations of the model itself, which relies on its inherent knowledge, whereas expert models are better equipped to generate effective steering vectors. 
More results are in \autoref{sec:appendix_source}.

\paragraph{It is essential for \ours to first extract features and subsequently align the representations.}
As shown in \autoref{fig:framework}, we first extract hidden-state features from the expert model, align them to the target models with trained auto-encoders, and then perform the intervention. We refer to this approach as \emph{RFMs-AE}. Alternatively, we can first align the sizes of hidden states using auto-encoders and then extract steering vectors by modifying \autoref{alg:RFMs} \autoref{line:krr} as follows:
\begin{equation}
    \pi^t(z) = \mathbb{K}^t(f_{\theta_i}(H_i),f_{\theta_i}(z)) \beta_t, \quad \text{where} \quad \beta_t = \left( \mathbb{K}^t(f_{\theta_i}(H_i), f_{\theta_i}(H_i)) \right)^{-1} Y
\end{equation}
This approach is referred to as \emph{AE-RFMs}. Experimental results in \autoref{fig:order-llama1b} show that \emph{RFMs-AE} consistently outperforms \emph{AE-RFMs}. This indicates that applying RFMs directly to raw hidden states preserves the integrity of the original feature space during the critical feature extraction phase, capturing nuanced patterns that might otherwise be lost with dimensionality reduction. This aligns with multimodal fusion research, which indicates that feature extraction prior to dimensionality reduction enhances performance \citep{DBLP:journals/pami/BaltrusaitisAM19}. By retaining original features during extraction, our approach generates more informative steering vectors for intervention.

\subsection{Discussion}
\label{sec:analysis_discussion}

\begin{wrapfigure}[14]{r}{9cm}
  \centering
    \begin{minipage}{0.57\linewidth}
      \centering
      \includegraphics[width=0.98\linewidth]{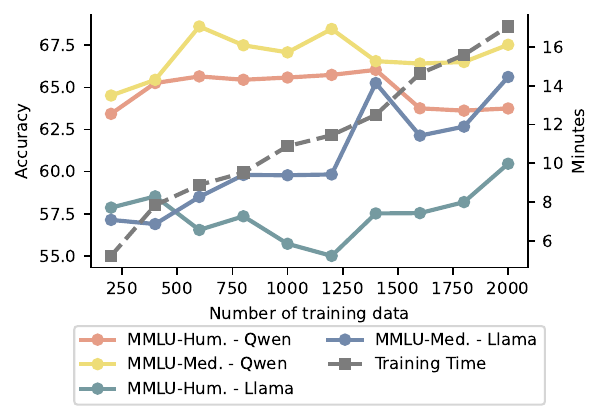}
      \caption{Training cost with \llamaE and \qwen as target model.}
      \label{fig:datacnt}
    \end{minipage}
    \hfill
    \begin{minipage}{0.4\linewidth}
      \centering
      \includegraphics[width=\linewidth]{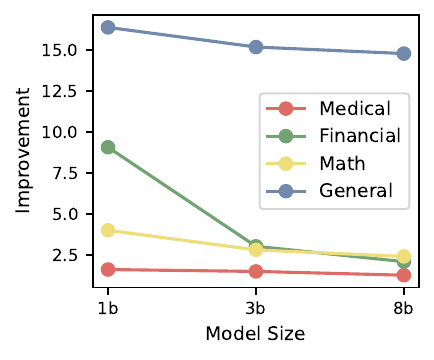}
      \caption{Performance gains of \ours across various model sizes on four domains.}
      \label{fig:scale}
    \end{minipage}
\end{wrapfigure}

\paragraph{\ours demonstrates high computational efficiency.}

As shown in \autoref{fig:datacnt}, increasing the training data volume linearly increases training time without necessarily improving accuracy on \mmlumed and \mmluchat tasks. We demonstrate that 2,000 training examples are sufficient for generating effective steering vectors, with an affordable time cost of approximately 17 minutes. Moreover, as detailed in \autoref{eq:intervention}, \ours adds only a single constant vector per layer. By adding $\varepsilon \cdot f_{\theta_i} (\nu_i)$ to the hidden states as a bias term, our intervention imposes negligible computational overhead during inference, highlighting the efficiency of our method, making it both scalable and practical.

\paragraph{\ours delivers larger performance gains with smaller models.}
We further investigate the effectiveness of \ours across varying model sizes. We conduct experiments with the Llama series (\llamaE, \llamaT, \llamaO) and present the results in \autoref{fig:scale}. \ours consistently improves performance across all the model sizes. Notably, we observe that \ours yields larger performance gains in smaller models. This trend can be attributed to the fact that smaller models have a limited capacity to store knowledge, making them benefit more from external interventions like \ours. .

% \begin{table}[] 
\begin{wraptable}[10]{t}{5.3cm}
\small
\centering
\setlength{\tabcolsep}{1pt}
\caption{
Results of \ours using base model as the target model.
}
\begin{tabular}{lcccc}
\toprule
& \begin{tabular}[c]{@{}c@{}}\dataset{MMLU}\\ \dataset{Med.} \end{tabular} & \begin{tabular}[c]{@{}c@{}}\dataset{MMLU}\\ \dataset{Fin.} \end{tabular} & \begin{tabular}[c]{@{}c@{}}\dataset{MMLU}\\ \dataset{Hum.} \end{tabular} & \begin{tabular}[c]{@{}c@{}}\dataset{MMLU}\\ \dataset{Math} \end{tabular} \\ \midrule
\multicolumn{5}{c}{\cellcolor{gray!15}\model{Llama-3.1-8B}}                           \\
Baseline & 25.11       & 26.64         & 25.22    & 24.33    \\
\ours    & \textbf{26.29}       & \textbf{30.69}         & \textbf{26.56}    & \textbf{26.39}    \\
\multicolumn{5}{c}{\cellcolor{gray!15}\model{Qwen2.5-7B}}                    \\
Baseline & 57.92       & 59.87         & 59.17    & 36.70    \\
\ours    & \textbf{59.14}       & \textbf{60.97 }        & \textbf{59.55}    & \textbf{38.28}   \\
\bottomrule
\end{tabular}
\label{tab:foundation}
% \end{table}
\end{wraptable}
\paragraph{\ours demonstrates effectiveness when using base models as the target models.}

Building on our earlier findings that \ours boosts performance, we now explore its impact on base models by applying it to \mmlu tasks with \model{Llama-3.1-8B} and \model{Qwen2.5-7B}. As shown in \autoref{tab:foundation}, although \ours again improves results, the gains are smaller than with SFT target models, referring to \autoref{tab:main}, because the steering vectors (derived from SFT expert models) face a larger distributional gap when applied to base models. This gap reduces effectiveness of the steering vectors in transferring expertise to the base models.

\section{Conclusion}
\label{sec:conclusion}

In this work, we introduce \ours, a novel activation steering method designed to enable knowledge transfer from any expert model to arbitrary target LLMs. Our approach consists of four key steps: (1) aligning the dimensionalities of the expert and target models using auto-encoders, (2) identifying optimal layer pairs for intervention through mutual information analysis, (3) generating steering vectors via Recursive Feature Machines (RFMs) from the identified expert layers, and (4) applying these steering vectors to the identified target layers. Results demonstrate that \ours significantly outperforms a wide range of baselines across diverse setups. This study advances the activation steering research in LLMs by introducing an effective and efficient intervention technique.
\section*{Acknowledgement}

This work is funded by EU Horizon Europe (HE) Research and Innovation programme grant No 101070631, and UK Research and Innovation under the UK HE funding grant No 10039436.

The computations described in this research were performed using the Baskerville Tier 2 HPC service (https://www.baskerville.ac.uk/). Baskerville was funded by the EPSRC and UKRI through the World Class Labs scheme (EP/T022221/1) and the Digital Research Infrastructure programme (EP/W032244/1) and is operated by Advanced Research Computing at the University of Birmingham.

\bibliographystyle{unsrt}
\bibliography{neurips_2025}

\clearpage
\appendix

\section{Details of Tasks}
\label{sec:appendix_tasks}
We list the detailed tasks in \mmlumed, \mmlufin, \mmlumath, and \mmluchat as follows: 
\begin{itemize}
    \item \mmlumed: It contains six tasks: \dataset{Anatomy}, \dataset{Clinical Knowledge}, \dataset{College Biology}, \dataset{College Medicine}, \dataset{Medical Genetics}, \dataset{Professional Medicine}.
    \item \mmlufin: It contains three tasks: \dataset{Econometrics}, \dataset{High School Macroeconomics}, \dataset{High School Microeconomics}.
    \item \mmlumath: It contains four tasks: \dataset{Abstract Algebra}, \dataset{College Mathematics}, \dataset{Elementary Mathematics}, \dataset{High School Mathematics}.
    \item \mmluchat: It contains twelve tasks: \dataset{Formal Logic}, \dataset{Global Facts}, \dataset{High School European History}, \dataset{High School US History}, \dataset{High School World History}, \dataset{Human Aging}, \dataset{Logical Fallacies}, \dataset{Moral Disputes}, \dataset{Moral Scenarios}, \dataset{Philosophy}, \dataset{Prehistory}, \dataset{World Religions}. 
\end{itemize}

To assess the safety of the LLMs, we follow \citep{llama} and evaluate the performance with a fine-tuned harmful classifier based on the DeBERTaV3.\footnote{\url{https://huggingface.co/domenicrosati/deberta-v3-xsmall-beavertails-harmful-qa-classifier}}Moreover, we use SacreBLEU to evaluate the performance on the \dataset{Flores-en2zh} and \dataset{Flores-zh2en} tasks.

\section{Baselines}
\label{sec:appendix_baselines}
To validate the effectiveness of our method, we select the followig methods as baselines:

\begin{itemize} 
  \item \textbf{Supervised Fine-Tuning (SFT):} We fine-tune all parameters of LLMs using the AdamW optimizer with a learning rate of $1 \times 10^{-5}$ and a batch size of 8. This process is conducted over three epochs on 2 NVIDIA A100 GPUs (80GB). During training, we use a linear learning rate schedule with a warm-up phase that constitutes 10\% of the total training steps. 
  \item \textbf{Knowledge Distillation (KD):} We use the expert model as the teacher and the LLMs as the student. The student model is trained on the instruction-tuning training set of each domain with the knowledge distillation loss. \cite{kd} proposes a method designed to facilitate knowledge distillation between teacher models and student models by leveraging optimal transport theory to enable distillation across models with different architectures and tokenizers. 
  \item \textbf{Inference-Time Intervention (ITI):} \cite{iti} operates by modifying the activations of specific attention heads during inference. ITI identifies a subset of attention heads within the model that exhibit high linear probing accuracy for the classification of positive answers and the corresponding negative answers. During inference, activations are shifted along directions calculated based on the linear probes. 
  \item \textbf{Contrastive Activation Addition (CAA):} \cite{caa} computes steering vectors by averaging the difference in the hidden states between pairs of positive and negative examples. During inference, these steering vectors are added at all token positions after the user's prompt with either a positive or negative coefficient, allowing precise control over the degree of the targeted behavior. 
  \item \textbf{Semantic-Adaptive Dynamic Intervention (SADI):} \cite{DBLP:journals/corr/abs-2410-12299} dynamically generates steering vectors tailored to each input's semantic context. SADI first computes activation differences between positive and negative pairs, which are then used to create a binary mask that highlights the most impactful model components. During inference, SADI applies the binary mask to the user input activations, scaling them element-wise based on the input's semantic direction, thereby dynamically steering the model's behavior. For ITI, CAA, and SADI, we extract steering vectors using the development set of each task to build the necessary contrastive pairs. 
\end{itemize}

\section{Training Details}
\label{sec:appendix_training}
We explore two knowledge-transfer scenarios. In the first, we transfer from a domain-specific expert model (e.g., medical, financial, mathematical) to a general-purpose target model by training auto-encoders on 2,000 domain-specific examples, using 500 domain-specific examples for mutual information analysis to identify intervention layers, and then employing 2,000 domain-specific examples as positive inputs alongside 2,000 general-domain examples as negative inputs to train RFMs. In the second scenario, we transfer from a larger general-purpose expert model to a smaller general-purpose target model. Similarly, we train the auto-encoders on 2,000 general-domain examples, using 500 general-domain examples to identify intervention layers. We then training RFMs on 2,000 general-domain examples as positive inputs and 2,000 domain-specific (e.g., medical) examples as negative inputs. All experiments are conducted on a single A100 GPU with 40 GB of memory.

\section{Hyperparameter Selection}
\label{sec:appendix-hyperparameter}
\begin{figure}[t]
  \centering
  \begin{minipage}{0.49\linewidth}
    \centering
    \includegraphics[width=\linewidth]{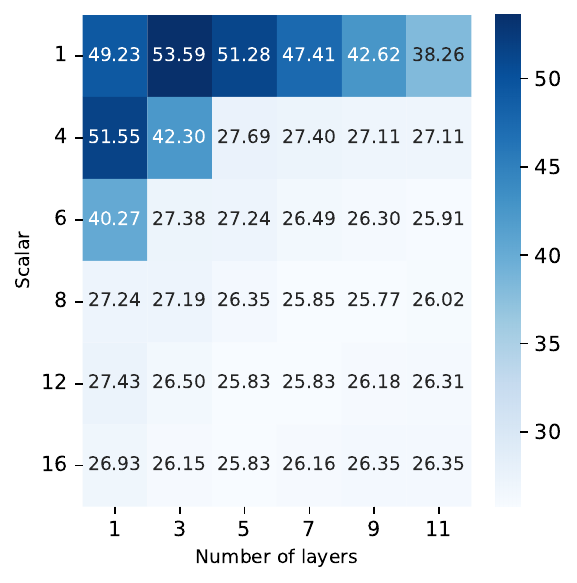}
    \caption{The selection of the number of intervention layers and scalar with \llamaE on the \medqa task.}
    \label{fig:hyper-llama8b}
  \end{minipage}
  \hfill
  \begin{minipage}{0.49\linewidth}
    \centering
    \includegraphics[width=\linewidth]{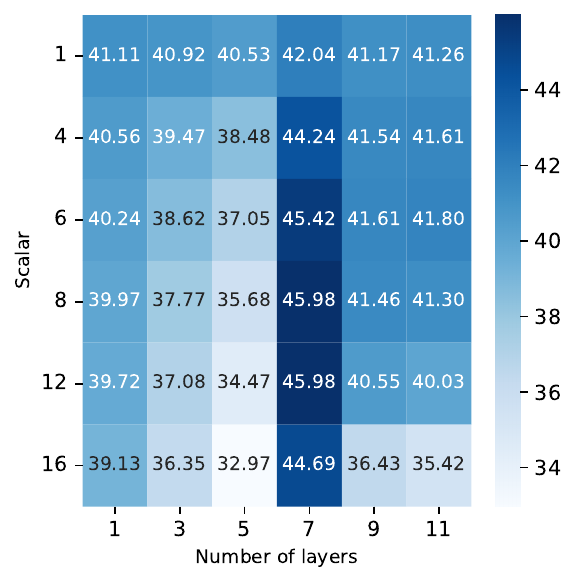}
    \caption{The selection of the number of intervention layers and scalar with \qwen on the \medqa task}
    \label{fig:hyper-qwen7b}
  \end{minipage}
\end{figure}

In \autoref{fig:hyper-llama8b} and \autoref{fig:hyper-qwen7b}, we sweep two hyperparameters to control the intervention: the number of intervention layers and the scalar. The number of intervention layers indicates how many layers we intervene in the model, and the scalar is used to control the strength of the intervention. Results indicate that the optimal settings for these hyperparameters vary across different models. This variability underscores that for precise task performance optimization, it is recommended to search for optimal hyperparameters using data from the validation sets with a small volume.% specific to each task.

\clearpage

\section{Supplementary Results of Feature Extraction Methods}
\label{sec:appendix_feature}
\begin{wraptable}[13]{t}{5.4cm}
\small
\centering
\setlength{\tabcolsep}{1pt}
\caption{
Comparison between different feature extraction methods on the medical tasks and general tasks.
}
\begin{tabular}{lcccc}
\toprule
& \dataset{MedQA}     & \begin{tabular}[c]{@{}c@{}}\dataset{MMLU}\\ \dataset{Med.} \end{tabular} & \copa     & \begin{tabular}[c]{@{}c@{}}\dataset{MMLU}\\ \dataset{Hum.}\end{tabular} \\ \midrule
\multicolumn{5}{c}{\cellcolor{gray!15}\gemma}                                     \\
Baseline & 28.63 & 31.81  & 72.32 & 34.36 \\
\ours \\
├ MD       & 28.69  & 32.06  &  72.83 & 34.38 \\
├ PCA      & 28.77  &  32.34 &  72.91 & 34.39 \\
└ RFMs  & \textbf{29.39} & \textbf{33.87}  & \textbf{75.57} &\textbf{34.63} \\
\bottomrule
\end{tabular}
\label{tab:feature-appendix1}
\end{wraptable}

We have demonstrated the effectiveness of \ours with RFM in \autoref{sec:analysis_ablation} using the \llamaE and \qwen backbones. In this section, we further validate the effectiveness of \ours with other feature extraction methods on the \gemma backbone. As shown in \autoref{tab:feature-appendix1}, \ours with RFMs outperforms PCA and MD across all tasks. This is consistent with the results in \autoref{sec:analysis_ablation}, which indicate that RFMs are more effective than simple linear feature extraction methods. Furthermore, we also provide comparisons on the \medmcqa, \nli, and \arcc tasks across three models in \autoref{tab:feature-appendix2}. The results show that \ours with RFMs consistently outperforms other feature extraction methods across all tasks.

\begin{table}
\small
\centering
\caption{
Comparison between different feature extraction methods on the medical tasks and general tasks.
}
\begin{tabular}{lccccccccc}
\toprule
 & \multicolumn{3}{c}{\llamaE}& \multicolumn{3}{c}{\qwen}& \multicolumn{3}{c}{\gemma}\\ \cmidrule(rl){2-4} \cmidrule(rl){5-7} \cmidrule(rl){8-10}
 &\begin{tabular}[c]{@{}c@{}} \dataset{Med}\\ \dataset{MCQA} \end{tabular} & \nli& \arcc&\begin{tabular}[c]{@{}c@{}} \dataset{Med}\\ \dataset{MCQA} \end{tabular}& \nli& \arcc&\begin{tabular}[c]{@{}c@{}} \dataset{Med}\\ \dataset{MCQA} \end{tabular}& \nli & \arcc \\ \midrule
Baseline&  49.40&  57.87&  67.39 &  46.25&  71.00&  73.54 &  33.06&  41.82& 38.17 \\
\ours \\
├ MD   &  48.89&  59.03&  67.49 &  47.43&  72.47&  73.69 &  33.11&  42.05& 38.31 \\
├ PCA  &  48.87&  59.08&  67.57 &  48.19&  73.94&  75.34 &  33.13&  42.33& 38.41 \\
└ RFM  &  \textbf{50.66}&  \textbf{61.36}&  \textbf{68.34} &  \textbf{48.57}&  \textbf{77.20}&  \textbf{78.24} &  \textbf{33.37}&  \textbf{44.10}& \textbf{39.11}\\ 
\bottomrule
\end{tabular}
\label{tab:feature-appendix2}
\end{table}

\section{Supplementary Results of Vector Generation Source}
\label{sec:appendix_source}

\begin{table}
\centering
\caption{Comparisons between steering vectors generated from model itself and expert model.}
\label{tab:vectorfrom-appendix}
\begin{tabular}{lccccccccc}
\toprule
 & \multicolumn{4}{c}{Medical}     & \multicolumn{5}{c}{NLU}   \\ \cmidrule(rl){2-5} \cmidrule(rl){6-10} 
\multicolumn{1}{c}{} & $\avgall$  & \dataset{MedQA} & \begin{tabular}[c]{@{}c@{}} \dataset{Med}\\ \dataset{MCQA} \end{tabular} & \begin{tabular}[c]{@{}c@{}}\dataset{MMLU}\\ \dataset{Med.} \end{tabular} & $\avgall$ &  \copa     &  \nli      &  \arcc     &  \begin{tabular}[c]{@{}l@{}}\dataset{MMLU}\\ \dataset{Hum.} \end{tabular}  \\
\midrule
\multicolumn{10}{c}{\cellcolor{gray!15}\textbf{\llamaE}} \\
 Baseline& 52.00& 45.60& 49.40& 60.99& 64.68& 74.01& 57.87& 67.39&59.45 \\
Self-generated&  53.60&  48.30&  49.51&  62.98&  65.29&  76.48&  57.81&  67.39& 59.49 \\
Expert-generated&  56.98&  53.59&  50.66&  66.71&  68.45&  83.47&  61.36&  68.34& 60.60 \\
\arrayrulecolor{gray!20}\midrule
\multicolumn{10}{c}{\cellcolor{gray!15}\textbf{\qwen}}  \\
Baseline&  49.65&  41.20&  46.25&  61.50&  72.51&  82.07&  71.00&  73.54& 63.41
\\
Self-generated&  50.08&  41.30&  46.74&  62.22&  78.52&  94.17&  80.12&  75.22& 64.60
\\
Expert-generated&  54.03&  45.98&  48.57&  67.53&  77.53&  88.23&  77.20&  78.24& 66.44
\\
\arrayrulecolor{gray!20}\midrule
\multicolumn{10}{c}{\cellcolor{gray!15}\textbf{\gemma}} \\
Baseline&  31.17&  28.63&  33.06&  31.81&  46.67&  72.32&  41.82&  38.17& 34.36
\\
Self-generated&  31.17&  28.64&  33.09&  31.79&  47.03&  73.41&  42.15&  38.22& 34.35
\\
Expert-generated&  32.21&  29.39&  33.37&  33.87&  48.35&  75.57&  44.10&  39.11& 34.63
\\
\arrayrulecolor{black}\bottomrule
\end{tabular}

\end{table}

The steering vectors used in previous studies are extracted from the model itself \cite{iti,caa}, but we argue that the steering vectors should be more effective if they are generated by expert models. In this section, we investigate the effectiveness of using steering vectors generated from the model itself (Self-generated) and those generated from expert models (Expert-generated). As shown in \autoref{tab:vectorfrom-appendix}, we find that the steering vectors generated from expert models are more effective than those generated from the model itself. This indicates that the steering vectors generated from expert models can better capture additional knowledge and improve the performance of \ours. These findings are

\begin{wraptable}[10]{t}{5cm}
\centering
\small
\setlength{\tabcolsep}{1pt}
\caption{
Comparisons of different kernels used in RFMs on the \llamaE. 
}
\begin{tabular}{lcccc}
\toprule
      Kernel   & $\avgall$  & \dataset{MedQA} & \begin{tabular}[c]{@{}c@{}} \dataset{Med}\\ \dataset{MCQA} \end{tabular} & \begin{tabular}[c]{@{}c@{}}\dataset{MMLU}\\ \dataset{Med.} \end{tabular} \\ \midrule
         baseline&  52.00&  45.60&  49.40& 60.99
\\ \midrule
         Laplacian &  56.98&  53.59&  50.66& 66.71
\\
         Gaussian &  53.86&  46.98&  50.83& 63.77
\\
         Linear &  53.41&  47.31&  49.55& 63.36
\\
\bottomrule
\end{tabular}
\label{tab:kernel-appendix}
\end{wraptable}
consistent with the results in \autoref{fig:modelA} in \autoref{sec:analysis_ablation}, which show that expert models provide more effective guidance for generation.

\section{Results of Other Kernels}
\label{sec:appendix_kernel}

As discussed in \autoref{sec:method_steering}, we implement RFMs with the Laplacian kernel. In this section, we further investigate the effectiveness of \ours with other kernels, including the Gaussian kernel and the Linear kernel. As shown in \autoref{tab:kernel-appendix}, we find that RFMs with the Laplacian kernel consistently outperforms other kernels across all tasks. This indicates that the Laplacian kernel is more effective in extracting the knowledge from the expert model, validating the effectiveness of our design choice.

\end{document}